\pdfoutput=1

\documentclass[sigconf,nonacm]{acmart}
\makeatletter\@ACM@balancefalse\makeatother
\usepackage{flushend}
\usepackage{balance}

\makeatletter\def\@titlefont{\fontsize{15}{19}\selectfont\sffamily\bfseries}\makeatother

\usepackage{graphicx}
\usepackage{pdflscape}
\usepackage{capt-of}
\usepackage{pdfpages}
\usepackage{amsmath}
\usepackage{etoolbox}
\usepackage{stfloats}
\patchcmd{\maketitle}{\setcounter{footnote}{1}}{\setcounter{footnote}{0}}{}{}
\patchcmd{\maketitle}{\setcounter{footnote}{1}}{\setcounter{footnote}{0}}{}{}
\usepackage{booktabs}
\usepackage{enumitem}
\usepackage{array}
\usepackage{tabularx}
\graphicspath{{figures/}}

\setcopyright{none}
\renewcommand\footnotetextcopyrightpermission[1]{}

\begin{document}
\clubpenalty=150 \widowpenalty=150

\title{ANASSA: From High-Level Framework to Low-Level Architecture for Spatial Agentic AI}

\author{Constantinos Papantoniou}
\authornote{Corresponding author: Constantinos Papantoniou $\cdot$ contact@anassa.ai. LinkedIn: \url{https://www.linkedin.com/in/constantinos-papantoniou/}}
\affiliation{%
  \institution{Anassa.ai}
  \city{}
  \country{USA}}

\author{Brian Hilton}
\authornote{LinkedIn: \url{https://www.linkedin.com/in/brian-hilton-ph-d/}}
\affiliation{%
  \institution{Research Scientist}
  \city{}
  \country{USA}}

\renewcommand{\shortauthors}{Papantoniou and Hilton}

\begin{abstract}
Agentic AI systems operating over geospatial data require more than autonomous reasoning and tool use; they must also support explicit orchestration, spatial validation, governance, provenance, failure handling, and human decision authority. Prior work introduced ANASSA, an agentic AI orchestration framework for spatial intelligence that established high-level components and responsibility boundaries for governed agentic execution~\cite{papantoniou2026anassa}. However, the framework did not specify how those high-level responsibilities should be systematically decomposed into an implementation-facing low-level architecture.

This paper addresses that gap by applying Design Science Research (DSR) to refine the ANASSA framework from high-level architectural (HLA) responsibilities into a low-level architecture (LLA). The resulting design preserves the responsibility boundaries of the foundational framework while translating them into executable architectural structures, including 47 logical roles, versioned task graphs, capability-based agent selection, typed contracts, explicit runtime states, spatial validation mechanisms, governance controls, failure and recovery paths, and provenance records.

The architecture is demonstrated through an end-to-end execution trace and analytically evaluated for requirements traceability, internal consistency, responsibility preservation, and technology independence. The resulting high-level-to-low-level design provides an implementation-facing bridge between the ANASSA framework and future system implementations, defining how governed spatial agentic workflows can be made inspectable, reconstructable, and testable without binding the architecture to a specific AI model, geospatial platform, communication protocol, or deployment environment.
\end{abstract}

\ccsdesc[500]{Information systems~Geographic information systems}
\ccsdesc[500]{Computing methodologies~Artificial intelligence}
\ccsdesc[500]{Computing methodologies~Multi-agent systems}

\keywords{geospatial agentic AI, spatial intelligence, design science research, software architecture, multi-agent orchestration, high-level architecture, low-level architecture, task graphs, typed contracts, spatial validation, provenance, governance}

\maketitle

\section{Introduction}

Geospatial operations increasingly depend on AI systems that can interpret objectives, reason over spatial and temporal context, invoke analytical capabilities, validate claims, and coordinate actions across heterogeneous data and execution environments. Recent geospatial-agent research demonstrates the value of structured workflows, specialized roles, tool invocation, retrieval, and feedback-driven execution, while also exposing persistent limitations in tool sequencing, spatial grounding, parameter inference, heterogeneous data handling, and long-chain robustness~\cite{wang2024survey,xi2023rise,hashemi2026survey,hu2025geoagent,bao2026spatialagent,bhattaram2025geoflow,liang2026geoagenticrag,yu2026geoagentbench}. These limitations are not only model limitations. They are system-design problems concerning how intent, reasoning, execution, evidence, governance, validation, recovery, and learning are connected.

Papantoniou and Hilton introduced ANASSA (Autonomous Neural Agents for Spatial Systems Architecture) as an agentic AI orchestration framework for spatial intelligence and established the framework-level architecture used as the starting point for this study~\cite{papantoniou2026anassa}. Rather than restating that specification, the present study treats its component boundaries, governance and evidence constraints, interface requirements, and evaluation limits as inherited design inputs, specifically, as high-level architectural (HLA) responsibilities. The new work begins where that baseline stops: at the systematic refinement of those responsibilities into an implementation-facing low-level architecture (LLA).

The remaining design problem is therefore not whether ANASSA should contain these capabilities, but how the high-level framework should be transformed into a coherent design solution that is precise enough to guide implementation while preserving its geospatial, governance, and traceability properties. A high-level architecture identifies responsibility boundaries and major relationships. An implementation-facing design must additionally specify logical roles, interface contracts, runtime states, capability resolution, failure semantics, validation paths, policy controls, and provenance records. It must also remain independent of a particular model provider, geospatial platform, cloud, or protocol so that the architecture does not collapse into a vendor-specific implementation.

This paper addresses that design problem using Design Science Research (DSR). DSR is appropriate when research advances knowledge through the purposeful creation and evaluation of artifacts that address a relevant problem~\cite{hevner2004design,peffers2007dsrm,gregor2013positioning}. Following the DSR process of problem identification, objectives for a solution, design and development, demonstration, evaluation, and communication~\cite{peffers2007dsrm}, this study treats the inherited ANASSA framework as kernel knowledge and a design input. The research artifact is the HLA-to-LLA refinement design: the responsibility interpretation, decomposition method, implementation-facing LLA, contracts, runtime semantics, and evaluation criteria that make the refinement explicit and testable.

The paper makes five contributions:
\begin{enumerate}[leftmargin=*,itemsep=2pt]
  \item It applies a DSR methodology to translate the foundational ANASSA framework into an implementation-facing design artifact with explicit problem, objectives, requirements, design principles, demonstration, and evaluation criteria.
  \item It derives functional and non-functional design requirements for governed geospatial agentic AI and formalizes eight design principles that constrain how ANASSA responsibilities are decomposed.
  \item It interprets the inherited four-layer C1--C11 structure as an HLA responsibility model, making ownership, governance and evidence boundaries, and implementation surfaces explicit without redefining the foundational framework.
  \item It specifies an implementation-facing LLA that refines the inherited architecture through logical-role decomposition, versioned task graphs, capability resolution, a canonical contract schema, observable runtime states, spatial validation, policy controls, bounded recovery, and provenance.
  \item It evaluates the design artifact through requirements-to-design traceability, internal consistency analysis, an end-to-end synthetic demonstration, and technology-independence analysis while reserving empirical performance evaluation for a subsequent reference implementation.
\end{enumerate}

The contribution is therefore the systematic refinement design and its implementation-facing specification, not a performance claim. The foundational ANASSA study defines the framework-level baseline~\cite{papantoniou2026anassa}; the present study explains how that baseline is refined into a coherent HLA-to-LLA system design, while subsequent work can instantiate and benchmark the design.

\section{Related Work and Architectural Positioning}

Recent geospatial-agent systems increasingly use structured workflows, specialized roles, and runtime tool execution, but they emphasize different parts of the autonomy problem. Spatial-Agent formalizes geospatial question answering as concept transformation into executable GeoFlow graphs and demonstrates the value of scientifically grounded DAG representations~\cite{bao2026spatialagent}. GeoFlow makes tool-calling objectives explicit inside generated geospatial workflows and reports improved execution success and token efficiency~\cite{bhattaram2025geoflow}. GeoAgentic-RAG integrates specialized retrieval, spatial-analysis, reasoning, and supervisory roles over vector, raster, and multimodal data~\cite{liang2026geoagenticrag}. GeoAgentBench shifts attention to dynamic execution by evaluating parameter selection, runtime feedback, and error recovery across multi-step geospatial tasks~\cite{yu2026geoagentbench}.

These systems motivate ANASSA but do not make the same architectural object the unit of contribution. The foundational ANASSA study focused on integrating reasoning, execution, spatial validation, governance, authoritative evidence, provenance, uncertainty, and human authority at the framework level~\cite{papantoniou2026anassa}. The present paper focuses on the implementation-facing execution architecture needed to preserve those properties across task decomposition, component boundaries, runtime failure, and replanning. Table~\ref{tab:related} summarizes this positioning; ``not a reported architectural focus'' does not imply that a system cannot support the capability through implementation choices.

\begin{table*}[t]
\caption{Positioning of the ANASSA LLA relative to representative geospatial-agent work.}
\label{tab:related}
\scriptsize
\begin{tabular}{p{0.13\textwidth}p{0.13\textwidth}p{0.11\textwidth}p{0.13\textwidth}p{0.16\textwidth}p{0.13\textwidth}}
\toprule
\textbf{System} & \textbf{Workflow representation} & \textbf{Multi-agent coordination} & \textbf{Runtime feedback / recovery} & \textbf{Spatial contract / evidence semantics} & \textbf{Governance / provenance as architecture} \\
\midrule
Spatial-Agent~\cite{bao2026spatialagent} & Scientific-concept GeoFlow DAG & Agent-centric; not multi-agent focus & Executable workflow generation & Spatial concepts and ordering constraints & Not a reported architectural focus \\
GeoFlow~\cite{bhattaram2025geoflow} & Agentic workflow with tool-calling objectives & Not a reported multi-agent focus & Runtime tool invocation & Tool objectives; no ANASSA-style cross-component contract & Not a reported architectural focus \\
GeoAgentic-RAG~\cite{liang2026geoagenticrag} & Multi-stage retrieval-analysis workflow & Specialized agents plus supervisor & Reflection/planning inside execution workflow & Unified geospatial database and retrieval context & Not a reported architectural focus \\
GeoAgentBench~\cite{yu2026geoagentbench} & Plan-and-React execution structure & Benchmark/agent architecture & Dynamic feedback and error recovery & Execution parameters and output verification & Benchmark focus rather than governance architecture \\
ANASSA HLA~\cite{papantoniou2026anassa} & Component task-graph requirement & C1--C11/C5 roles & Typed failure requirement & Authoritative evidence, uncertainty, provenance & Explicit C7 governance and human authority \\
\textbf{ANASSA LLA (this paper)} & \textbf{Versioned task DAG per revision} & \textbf{47 logical roles + C4 orchestration} & \textbf{Typed states, retry, replan, escalation} & \textbf{Typed spatial contracts + evidence states} & \textbf{Runtime policy, audit, provenance, authority} \\
\bottomrule
\end{tabular}
\end{table*}

From a DSR perspective, these works provide important kernel knowledge for task graphs, tool-augmented reasoning, specialist coordination, runtime feedback, and recovery. The ANASSA contribution is positioned at the architecture-design level: it assembles these capability classes with explicit geospatial evidence, governance, policy, provenance, and deployment boundaries into a design artifact whose elements can later be instantiated and evaluated. This positioning follows DSR guidance that the artifact and the knowledge embodied in its design should be made explicit rather than presented as an undifferentiated implementation~\cite{hevner2004design,gregor2013positioning}.

\section{Design Science Research Methodology}

\subsection{DSR Framing and Research Artifact}

Design Science Research focuses on the purposeful creation and evaluation of artifacts that address relevant problems and contribute reusable design knowledge~\cite{hevner2004design}. This paper follows the six-activity DSR process described by Peffers et al.: problem identification and motivation, definition of objectives for a solution, design and development, demonstration, evaluation, and communication~\cite{peffers2007dsrm}. Gregor and Hevner further emphasize the need to make the artifact, its level of abstraction, and its knowledge contribution explicit~\cite{gregor2013positioning}.

The research artifact in this study is the \textbf{ANASSA HLA-to-LLA refinement design}. The foundational ANASSA framework serves as kernel knowledge and a design input rather than as a new contribution. The artifact consists of: (1) functional and non-functional requirements for implementation-facing refinement; (2) eight LLA-focused design principles; (3) an HLA responsibility interpretation; (4) the HLA-to-LLA decomposition method; (5) the LLA role, task, contract, state, validation, governance, recovery, and provenance models; and (6) traceability and evaluation criteria that make the refinement testable in a later implementation.

Table~\ref{tab:dsrmap} maps the DSR activities to this paper.

\begin{table*}[t]
\caption{Application of Design Science Research Methodology to the ANASSA architecture design.}
\label{tab:dsrmap}
\small
\begin{tabular}{p{0.19\textwidth}p{0.33\textwidth}p{0.39\textwidth}}
\toprule
\textbf{DSR activity} & \textbf{ANASSA design activity} & \textbf{Artifact / evidence in this paper} \\
\midrule
Problem identification and motivation & Identify the gap between capable geospatial agents and an implementation-ready governed architecture & Problem statement, foundational ANASSA framework, related-work comparison \\
Objectives for a solution & Define what the architecture must enable and constrain & Design objectives, functional requirements, non-functional requirements, inherited DR1--DR8 \\
Design and development & Translate requirements into principles, HLA responsibility boundaries, and LLA execution mechanisms & DP1--DP8, Figure~\ref{fig:hla}, Figure~\ref{fig:lla}, task graphs, contracts, runtime states, validation and recovery \\
Demonstration & Walk a representative geospatial objective through the designed artifact & Synthetic end-to-end trace with a CRS mismatch, graph revision, policy decision, spatial validation, and provenance closure \\
Evaluation & Assess whether the design is complete and internally coherent at the architecture level & Requirements traceability, flow consistency, failure-path closure, policy/evidence separation, technology-independence analysis \\
Communication & Present the artifact and its derivation so it can be implemented and empirically tested & HLA/LLA diagrams, design tables, logical interaction patterns, evaluation variables, implementation-readiness criteria \\
\bottomrule
\end{tabular}
\end{table*}

\subsection{Problem Identification and Motivation}

The design problem is fragmentation across reasoning, geospatial data, tool execution, governance, authoritative evidence, and organizational decision authority. A model can produce a plausible plan while still losing coordinate reference information, relying on stale evidence, selecting an unavailable capability, violating policy, or failing without a reconstructable recovery path. The architecture must therefore do more than coordinate agents: it must carry the context and controls required to move safely from intent to validated action.

The foundational ANASSA study provides the starting architecture and eight design requirements~\cite{papantoniou2026anassa}. This paper treats those requirements as inherited constraints and refines them through a product/system design lens: who or what is responsible, what information crosses each boundary, what can fail, how the system recovers, what must be observable, and which responsibilities must remain replaceable across technology choices.

\subsection{Objectives for the Design Solution}

Five design objectives guide the artifact:
\begin{itemize}[leftmargin=*,itemsep=2pt]
  \item \textbf{DO1 -- Implementation precision:} define responsibility and interface boundaries precisely enough to guide a reference implementation.
  \item \textbf{DO2 -- Observable orchestration:} make task construction, capability selection, state transition, failure, retry, replanning, policy decisions, validation, and escalation observable.
  \item \textbf{DO3 -- Geospatial integrity:} preserve spatial reference, geometry, temporal context, source authority, quality, and uncertainty across component boundaries.
  \item \textbf{DO4 -- Governed autonomy:} allow machine-paced operation within policy while preserving explicit human authority where risk, consequence, uncertainty, or regulation requires it.
  \item \textbf{DO5 -- Deployment independence and reproducibility:} separate logical capabilities from specific model, platform, cloud, and protocol implementations while retaining sufficient provenance to reconstruct a completed cycle.
\end{itemize}

\subsection{Functional and Non-Functional Design Requirements}

The functional requirements describe what ANASSA must do; the non-functional requirements describe the system qualities the design must preserve. They derive from the foundational ANASSA requirements, failure modes identified in related geospatial-agent work, and the design objectives above.

\begin{table*}[t]
\caption{Functional and non-functional requirements for the ANASSA design solution.}
\label{tab:frnfr}
\scriptsize
\begin{tabular}{p{0.05\textwidth}p{0.39\textwidth}p{0.05\textwidth}p{0.39\textwidth}}
\toprule
\textbf{ID} & \textbf{Functional requirement} & \textbf{ID} & \textbf{Non-functional requirement} \\
\midrule
FR1 & Translate participant objectives into machine-actionable, policy-aware geospatial intent. & NFR1 & Traceability: reconstruct data, task, model/tool, policy, validation, and human-decision paths. \\
FR2 & Perform geospatial reasoning and select appropriate analytical capabilities and workflows. & NFR2 & Reproducibility: retain versions, parameters, graph revisions, and evidence identifiers for repeated execution. \\
FR3 & Orchestrate tool and service execution across heterogeneous platforms and deployment environments. & NFR3 & Interoperability: exchange typed information across replaceable geospatial tools and services. \\
FR4 & Validate spatial claims and artifacts against governed authoritative evidence and explicit quality conditions. & NFR4 & Deployment independence: avoid binding logical orchestration to one model, platform, cloud, or protocol. \\
FR5 & Apply governance, authorization, risk, privacy, sovereignty, and human-authority controls at runtime. & NFR5 & Extensibility: add sectors, capabilities, agents, tools, or evidence sources without redesigning the full architecture. \\
FR6 & Detect execution and validation failures and select bounded retry, replanning, escalation, or termination. & NFR6 & Scalability: separate orchestration semantics from workload size, data volume, workflow depth, and distributed compute. \\
FR7 & Capture permitted learning and adaptation from completed cycles without allowing autonomous modification of protected controls. & NFR7 & Security and sovereignty: preserve access, credential, data-residency, and jurisdictional constraints. \\
FR8 & Deliver validated geospatial intelligence together with decision, evidence, validation, and provenance records. & NFR8 & Auditability: expose machine-readable states and control decisions sufficient for review and accountability. \\
\bottomrule
\end{tabular}
\end{table*}

\subsection{Design and Development Procedure}

The architecture is developed through constrained decomposition rather than by enumerating agents first. \textbf{Step 1: freeze inherited invariants.} The four layers, C1--C11 boundaries, C5 sequence, C7/C8 authority relationships, and DR1--DR8 are treated as fixed inputs. \textbf{Step 2: assign product/system responsibilities.} Each HLA component is defined by the responsibility it owns and by the responsibilities it must not absorb. \textbf{Step 3: decompose responsibilities into capabilities and logical roles.} A role is created only when it has a distinguishable responsibility, contract, policy scope, or observable state. \textbf{Step 4: define interfaces before implementations.} Cross-boundary exchanges receive typed payload, spatial/temporal context, provenance, policy, validation, and failure semantics. \textbf{Step 5: define runtime behavior.} Task-graph revision, capability resolution, state transitions, policy controls, validation, recovery, and provenance are specified. \textbf{Step 6: evaluate traceability and closure.} Every requirement must map to identifiable HLA/LLA elements, and an end-to-end cycle must be reconstructable without relying on hidden model reasoning.

This procedure is intentionally technology-agnostic. References to language or multimodal models, retrieval, vector databases, APIs, agent-to-agent protocols, geospatial software, and cloud infrastructure identify capability classes rather than required vendors. A logical role may be implemented by an AI agent, deterministic service, workflow engine, human-controlled function, or combination of these as long as the defined responsibility and observable records are preserved.

\subsection{Demonstration and Evaluation Strategy}

The artifact is demonstrated using a synthetic geospatial workflow that intentionally encounters a coordinate-reference mismatch and then exercises graph revision, policy evaluation, evidence-based validation, uncertainty retention, output packaging, and provenance closure. The evaluation is analytical rather than performance-based. Four criteria are used: \textbf{E1 requirements coverage}, \textbf{E2 internal consistency}, \textbf{E3 end-to-end design closure}, and \textbf{E4 technology independence}. Empirical accuracy, latency, cost, and scalability remain evaluation targets for a subsequent reference implementation.

\section{ANASSA Design Principles}

The design principles translate the objectives and requirements into rules that constrain HLA-to-LLA refinement. They operationalize and extend the framework-level principles established in the foundational ANASSA study~\cite{papantoniou2026anassa}; they are not presented as a replacement for that earlier set. The principles are intended to remain stable even when specific tools, models, runtimes, or deployment technologies change.

\begin{table*}[t]
\caption{ANASSA design principles used to guide HLA-to-LLA refinement.}
\label{tab:principles}
\scriptsize
\begin{tabular}{p{0.07\textwidth}p{0.24\textwidth}p{0.61\textwidth}}
\toprule
\textbf{ID} & \textbf{Design principle} & \textbf{Architectural implication} \\
\midrule
DP1 & Geospatial context is a first-class system property & Tasks, contracts, evidence, and outputs preserve coordinate reference, geometry, scale/resolution, temporal context, jurisdiction, quality, and uncertainty rather than treating them as hidden prompt context. \\
DP2 & Separate intent from execution & Participant intent is normalized and decomposed before tools are invoked; intent, planning, capability selection, and execution remain distinguishable and auditable. \\
DP3 & Orchestrate capabilities, not vendors & C4 resolves logical capabilities to replaceable implementations so the architecture is independent of a particular model provider, geospatial platform, cloud, or protocol. \\
DP4 & Use explicit contracts at every architectural boundary & Inter-component exchanges declare source, destination, payload, spatial/temporal context, provenance, constraints, policy, validation, expected result, and failure semantics. \\
DP5 & Validation is part of execution & Spatial reference, topology, positional accuracy, resolution, temporal validity, authority, uncertainty, and evidence applicability are evaluated as part of the runtime path rather than only after generation. \\
DP6 & Autonomy is policy- and risk-adaptive & Human authority is invoked according to risk, policy, consequence, uncertainty, and evidence state; low-risk operations may proceed within approved boundaries. \\
DP7 & Fail explicitly and recover from evidence & Failures are typed, recorded, and routed to bounded retry, graph revision, escalation, compensating action, or termination rather than hidden free-form retries. \\
DP8 & Every released decision must be reconstructable & A completed cycle preserves sufficient data, graph, tool/model, parameter, policy, validation, failure, evidence, and human-decision records to explain and reproduce the path to release. \\
\bottomrule
\end{tabular}
\end{table*}

The principles are not independent slogans. DP1 and DP5 preserve domain integrity; DP2--DP4 define the control and interface structure; DP6 constrains autonomy; DP7 defines failure behavior; and DP8 closes the system around accountability and reproducibility. Together they provide the design logic connecting the DSR objectives to the two architecture levels presented next.

\subsection{Requirements-to-Design Traceability}

Table~\ref{tab:traceability} connects the inherited ANASSA requirements to design principles, HLA responsibility owners, LLA mechanisms, and observable evidence. This traceability is the primary analytical check that the LLA remains a refinement of the HLA rather than an unrelated implementation design.

\begin{table*}[t]
\caption{Traceability from inherited ANASSA requirements through design principles to HLA/LLA mechanisms.}
\label{tab:traceability}
\scriptsize
\begin{tabular}{p{0.04\textwidth}p{0.14\textwidth}p{0.08\textwidth}p{0.10\textwidth}p{0.25\textwidth}p{0.23\textwidth}}
\toprule
\textbf{Req.} & \textbf{Requirement} & \textbf{DP} & \textbf{HLA owner} & \textbf{LLA mechanism} & \textbf{Observable evidence} \\
\midrule
DR1 & Spatial grounding & DP1, DP5 & C6, C8, C5-S4 & Spatial/temporal contract fields; preparation and validation checks & CRS/datum, geometry, resolution, time, transformation chain, validation record \\
DR2 & Execution-aware orchestration & DP2, DP3, DP7 & C4, C5 & Versioned task graph, capability registry, preconditions, typed runtime state & Graph revision, capability decision, tool/parameter record, failure stage \\
DR3 & Authoritative validation & DP4, DP5 & C8, C5-S4, C7 & Claim/artifact validation against governed evidence & Evidence bundle, source/version/authority, tolerance, verdict \\
DR4 & Uncertainty representation & DP1, DP5 & C6, C8, C5-S4 & Non-binary evidence states and uncertainty fields & Validated-with-uncertainty, conflict, insufficient-evidence state \\
DR5 & Human decision authority & DP6 & C7, C5-S3/S4 & Policy evaluation and risk-adaptive human control & Permit/deny/conditions or explicit human-authorization record \\
DR6 & Traceability and provenance & DP4, DP8 & C10, C11 & Cycle-level provenance linked to contracts, evidence, and graph revisions & Reconstructable data/model/tool/policy/human decision path \\
DR7 & Reproducible interfaces & DP3, DP4, DP7 & C4, C6, C7, C8 & Canonical contracts, logical interaction patterns, typed failures & Versioned schema, acceptance/rejection, failure/recovery record \\
DR8 & Scalable distributed execution & DP3 & C3, C4 & Separation of orchestration from transport/runtime implementation & Stable logical task/contract semantics across alternate runtimes and targets \\
\bottomrule
\end{tabular}
\end{table*}

\section{ANASSA HLA Responsibility Model}

\subsection{Four Layers and Eleven Components}

Figure~\ref{fig:hla} uses the component identifiers and four-layer boundary model inherited from the foundational ANASSA framework~\cite{papantoniou2026anassa}. The present study interprets this structure as an HLA responsibility map rather than re-specifying it component-by-component. For design purposes, the responsibilities form four ownership classes: external context and operating surfaces (C1--C3 and C6), orchestration and cognitive execution (C4--C5), governance and evidence authority (C7--C8), and process/output closure (C9--C11). This grouping makes the subsequent decomposition explicit without changing the inherited component boundaries.

The L0--L4 notation inside C4 is intentionally separate from the four top-level ANASSA layers. L0--L4 describe the internal execution stack of one HLA component; they do not introduce additional framework layers. This distinction is important because the HLA remains stable while implementation detail can evolve beneath C4.

\subsection{From Component Boundaries to Implementation Surfaces}

The HLA is intentionally compact. For implementation, each component boundary becomes one or more surfaces that must be made explicit. A surface may carry intent, data, model context, a policy decision, an execution request, a validation result, a human decision, a learning update, or a final intelligence artifact. The LLA therefore expands each component in three directions: \emph{role decomposition} (which agent or service is responsible), \emph{interface definition} (what is exchanged), and \emph{runtime semantics} (what states and controls govern the exchange).

This expansion preserves the original architecture while avoiding a common failure mode in agentic systems: allowing the foundation model to become the implicit architecture. In ANASSA, models are replaceable substrates inside explicit control, data, and validation boundaries.

\section{HLA-to-LLA Design Decomposition}

The HLA and LLA are not independent diagrams. The LLA is derived from the HLA through a repeatable refinement sequence:

\begin{equation}
\begin{aligned}
\text{HLA Responsibility} &\rightarrow \text{Capabilities} \rightarrow \text{Logical Roles}\\
&\rightarrow \text{Contracts} \rightarrow \text{Runtime Behavior}.
\end{aligned}
\end{equation}

For each HLA component, the design asks five questions: \textit{What outcome or control does the component own? What capabilities are required to fulfill that responsibility? Which responsibilities require separate logical roles? What information must cross the boundary? What states, policies, validation, and failures must be observable at runtime?} This sequence prevents the design from beginning with an arbitrary agent count or vendor stack.

C4 illustrates the method. At the HLA, C4 owns orchestration across the distributed geospatial system-of-systems. That responsibility decomposes into intent/task decomposition, capability resolution, sequencing, scheduling, policy enforcement, execution monitoring, recovery, and output synthesis. Those capabilities are organized into the L0 Orchestrator and L1--L4 supporting layers. The same method is applied to C1--C3, C5--C8, and C10, while C9/C11 retain their process-flow responsibility.

Three design rules control the refinement. First, \textbf{responsibility continuity}: every LLA role must trace to an HLA owner. Second, \textbf{boundary preservation}: decomposition may refine a component but may not silently move governance, evidence authority, or human authority into an unrelated component. Third, \textbf{implementation substitutability}: LLA roles and contracts describe what must be provided, while concrete models, geospatial tools, services, and runtimes remain replaceable.

Refinement is considered complete only when four architecture-review gates are satisfied. \textbf{Ownership closure} requires every logical role and state transition to have an HLA owner. \textbf{Interface closure} requires every cross-boundary dependency to be represented by an explicit contract rather than hidden prompt context or undocumented coupling. \textbf{Control closure} requires policy decisions, validation results, failure states, and escalation paths to be observable at the same boundary where the corresponding action occurs. \textbf{Substitution closure} requires a provider, model, tool, or runtime to be replaceable without changing the upstream logical task and contract semantics. These gates turn the HLA-to-LLA transformation into a reviewable design procedure rather than a diagramming exercise.

The output of the decomposition is therefore a consistent design package for each HLA owner: a responsibility statement, capability inventory, logical-role assignment, contract boundary, runtime-state and failure behavior, and evidence/provenance obligation. A component is not considered implementation-ready when any of these elements is missing. The same closure criteria are reused in Section~\ref{sec:dsreval} as part of the DSR artifact evaluation, connecting the design procedure directly to its evaluation rather than assessing the architecture after the fact.

During architecture review, a failed closure gate returns the component to refinement. A pass requires inspectable evidence: an owning component, role mapping, contract, state transition, control hook, bounded failure route, and provenance link. This makes the transformation auditable and prevents operational dependencies from being hidden in prompts or adapters. In product-design terms, HLA approval establishes ownership; LLA approval establishes executable, observable, governable, and replaceable boundaries.

\clearpage
\onecolumn
\refstepcounter{figure}\label{fig:hla}
\includepdf[pages=1,fitpaper=true,pagecommand={\thispagestyle{empty}}]{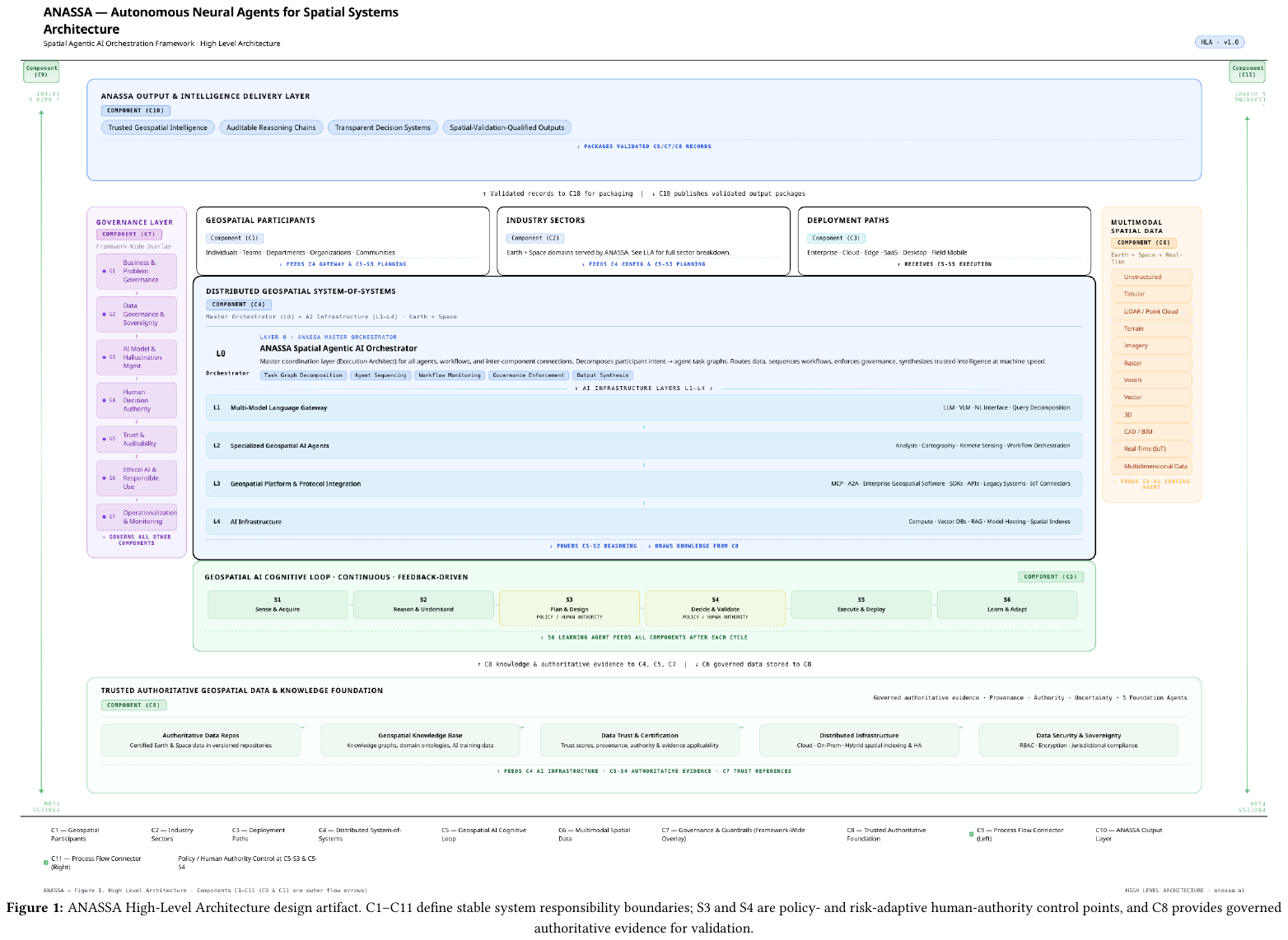}
\clearpage
\twocolumn

\section{Implementation-Facing Low-Level Architecture (LLA)}

\subsection{LLA Overview}

Figure~\ref{fig:lla} summarizes the low-level design artifact. Its decomposition is constrained by DP1--DP8 and the HLA-to-LLA derivation rules in the preceding section. The LLA therefore refines component responsibilities without redefining the HLA: it makes roles, contracts, task structures, states, policy checks, validation, recovery, and provenance explicit at the level required for implementation.

\subsection{C1--C3: Intent, Domain Context, and Deployment}

C1 converts human or organizational participation into machine-actionable intent while retaining identity, authorization, role, and collaboration context. Its logical roles include Participant Engagement, Collaborative Workflow, UX and Access, Community Mapping, and Multi-Stakeholder Coordination. These roles do not perform the full spatial analysis; they establish who is asking, what objective is being pursued, what data exposure is permitted, and which competing stakeholder constraints must be carried into planning.

C2 binds the workflow to sector-specific semantics. The LLA distinguishes Earth and space sectors and assigns roles for Industry Requirements, Regulatory and Compliance, Industry Workflow Automation, Domain Knowledge and Insight, and Sector Reporting and Communications. The important architectural property is not the enumerated sector list itself but the existence of a domain-context interface. Planning at S3 must be able to retrieve the applicable standards, tolerances, regulatory constraints, and output expectations without embedding them as untraceable prompt assumptions.

C3 represents deployment environments---enterprise, cloud, edge, hosted-service integration, desktop, and field/mobile---rather than a product classification of ANASSA. It assigns roles for Solution Recommendation, Deployment Monitoring and Optimization, Scalability and Performance, and Legacy Migration and Modernization. C3 receives approved execution from S5 and provides the runtime environment in which geospatial services, APIs, databases, models, and external operational systems are invoked.

\subsection{C4: Distributed Geospatial System-of-Systems}

C4 is the execution backbone of the low-level architecture. Its internal Layer 0 is the ANASSA Spatial Agentic AI Orchestrator, while L1--L4 provide the model, specialist-agent, integration, and infrastructure capabilities through which tasks are executed. The Orchestrator serves as the central orchestration control plane rather than as a message router. It owns the lifecycle of the workflow graph and maintains the coordination state that connects participant intent to evidence-bound output.

Layer 0 contains six core responsibility classes represented in Figure~\ref{fig:lla}: task-graph decomposition, agent sequencing, workflow monitoring, governance enforcement, output synthesis, and hallucination/validation escalation. An implementation may realize these functions as separate services or as modules inside one orchestration service, but their state transitions must remain distinguishable.

The supporting C4 stack is organized as follows:

\begin{itemize}[leftmargin=*,itemsep=2pt]
  \item \textbf{L1 -- Multi-Model Language Gateway:} language and multimodal model access, natural-language interface, query decomposition, prompt/instruction construction, and response synthesis.
  \item \textbf{L2 -- Specialized Geospatial AI Agents:} spatial analysis, cartography, remote sensing, workflow orchestration, and system-connectivity capabilities.
  \item \textbf{L3 -- Geospatial Platform Integration:} model/tool interoperability interfaces such as Model Context Protocol (MCP), agent-to-agent interfaces, enterprise geospatial software, SDKs, REST APIs, IoT connectors, legacy adapters, and OGC-aligned services.
  \item \textbf{L4 -- AI Infrastructure:} compute, model hosting/registry, vector and spatial indexes, retrieval infrastructure, caching, and supporting runtime services.
\end{itemize}

The L0--L4 decomposition separates orchestration policy from vendor implementation. A task may be reasoned over by one model, executed by a geospatial service from another vendor, retrieve evidence from an enterprise spatial database, and publish through a third platform without changing the top-level ANASSA component model.

\subsection{C6: Multimodal Spatial Data}

C6 is the preparation and streaming boundary between heterogeneous spatial observations and the Cognitive Loop. The LLA groups inputs into Earth data, space/orbital data, and real-time streams. These include vector and raster data, terrain, LiDAR and point clouds, imagery, voxel and 3D data, CAD/BIM, multidimensional cubes, orthorectified imagery, SAR, multispectral and hyperspectral data, orbital telemetry, space weather, IoT networks, traffic, GNSS, volunteered geographic information, drone and vehicle telemetry, emergency feeds, and industrial sensors.

Logical roles cover ingestion, quality and validation, ETL and transformation, streaming, spatial data fusion, metadata and catalog management, classification and labeling, and compression and optimization. The LLA requires C6 to emit prepared bundles with explicit spatial reference, time, source identifiers, quality metadata, and provenance. Missing coordinate systems, stale observations, incompatible resolution, or insufficient metadata are therefore visible before S1 rather than silently inferred downstream.

\subsection{C5: Geospatial AI Cognitive Loop}

The LLA does not redefine the six C5 stages established in the foundational ANASSA framework; it gives those stages implementation-facing exchange semantics~\cite{papantoniou2026anassa}. S1 emits an observation/event record, S2 a structured interpretation with selected capabilities, S3 a versioned task-graph proposal, S4 a validation and policy disposition, S5 an execution record bound to the approved graph, and S6 a governed learning update or change proposal. These records convert the inherited cognitive cycle into explicit interfaces that can be validated, replayed, and audited.

S6 (Learn and Adapt) produces bounded learning updates for permitted subscribers. S6 may update approved preference, capability, performance, and coordination state, but it may not autonomously modify C7 governance policy, C8 certification criteria, authorization privileges, or other protected control state; changes to those controls are emitted as proposals requiring authorized approval.

S3 and S4 are explicit human-authority control points. The LLA does not require a human to block every machine-paced transaction. Instead, C7 evaluates the workflow's risk class, policy, consequence, uncertainty, and evidence state to determine whether execution is auto-permitted, conditionally permitted, or held for explicit human authorization. Human decisions become provenance-bearing events rather than off-system approvals.

\subsection{C7, C8, C9--C11: Governance, Evidence, Process Flow, and Output}

The foundational ANASSA framework defines C7 governance and C8 evidence responsibilities; the LLA refines them into runtime control surfaces rather than repeating their high-level definitions~\cite{papantoniou2026anassa}. C7 exposes policy evaluation, authorization, audit, escalation, and monitoring decisions at scheduling and validation boundaries. The responsible logical roles produce machine-readable dispositions that can permit, deny, condition, or hold an action, with each decision linked to the task and provenance record. This treatment is consistent with lifecycle-oriented AI governance in which problem definition, data quality, model behavior, human oversight, and operational monitoring are connected controls~\cite{cognilytica2019cpmai}.

C8 exposes evidence retrieval, source certification metadata, authority/applicability checks, and security/sovereignty controls. The LLA requires evidence responses to carry jurisdiction, temporal validity, resolution or accuracy support, provenance, and applicable uncertainty rather than returning an unqualified ``ground truth'' value. S4 can therefore distinguish supported, uncertain, conflicting, insufficient, and rejected evidence conditions using the same contract structure~\cite{iso19157}.

C9 and C11 remain process-flow connectors. C9 carries data and intent downward into the architecture, while C11 returns learning, validation, and release state. C10 assembles the externally consumable release package from approved intelligence artifacts, machine-readable decision/provenance records, and the validation status attached to each spatial claim. C10 does not independently re-validate claims; it publishes results whose validation and governance dispositions are already explicit.

\section{Logical Roles and Orchestration Semantics}

\subsection{Five-Tier Logical Role Hierarchy}

The LLA inventories 47 logical roles across five functional tiers (Table~\ref{tab:hierarchy}). The functional hierarchy identifies ownership of orchestration, cognitive-cycle behavior, domain work, governance, and foundation services.

\begin{table}[t]
\caption{Five-tier ANASSA logical role hierarchy.}
\label{tab:hierarchy}
\small
\begin{tabularx}{\columnwidth}{c >{\raggedright\arraybackslash}X c >{\raggedright\arraybackslash}X}
\toprule
\textbf{Tier} & \textbf{Name} & \textbf{Count} & \textbf{Primary responsibility} \\
\midrule
A0 & ANASSA Spatial Agentic AI Orchestrator & 1 & Global workflow and connection coordination \\
A1 & Cognitive Loop Agents & 6 & Sense, reason, plan, decide/validate, execute, and learn \\
A2 & Domain Specialist Agents & 27 & Task-specific operational and geospatial capabilities \\
A3 & Governance Agents & 8 & Policy, audit, validation, oversight, and monitoring \\
A4 & Foundation Agents & 5 & Data, knowledge, trust, infrastructure, and security \\
\midrule
& \textbf{Total} & \textbf{47} & \\
\bottomrule
\end{tabularx}
\end{table}

The hierarchy is deliberately compatible with heterogeneous runtimes. A ``logical agent'' is defined by a responsibility, input/output contract, policy scope, and observable state. It need not correspond one-to-one with a process, container, model instance, or vendor agent object. This prevents implementation details from collapsing the architecture into a particular software framework.

\subsection{Task-Graph Representation}

Participant intent is not dispatched directly to a tool. C4 first translates the objective into a versioned executable task graph. For graph revision $r$, ANASSA represents the workflow as
\begin{equation}
G_r=(V_r,E_r),
\end{equation}
where each node $v_i\in V_r$ is a typed task and each edge $e_{ij}\in E_r$ is a declared dependency. A single graph revision is acyclic so that scheduling and provenance are well defined; the higher-level C5 Cognitive Loop is cyclic because failure, validation, and learning may produce a new graph revision $G_{r+1}$.

A canonical task node is represented logically as
\begin{equation}
\begin{aligned}
v_i=\langle & id,role,capability,input,space,time,\\
              & policy,validation,recovery,output \rangle .
\end{aligned}
\end{equation}
The fields carry the task/cycle identifier, requesting/executing role, required capability, input contract, spatial and temporal context, governance policy, validation rules, retry/replan behavior, expected output, and provenance pointer. Edges are typed as \emph{data}, \emph{control}, \emph{validation}, or \emph{governance} dependencies. Parallel branches are permitted only when their dependencies and policy/evidence constraints are satisfied.

Replanning never overwrites the prior graph. Instead, an observed runtime condition generates a new graph revision with a recorded reason and lineage to the superseded node(s). This preserves the difference between an adaptive workflow and an untraceable sequence of model-generated actions. The representation extends graph-based geospatial-agent work~\cite{bao2026spatialagent,bhattaram2025geoflow} by binding execution structure to spatial contracts, governance, evidence, typed runtime state, and provenance.

\subsection{Capability Resolution and Agent Selection}

For each task node, the Orchestrator resolves a capability rather than selecting an agent by name alone. The registry entry for a capability records the available implementation(s), supported geometry/data types, coordinate/reference constraints, and input/output schemas.

\clearpage
\onecolumn
\refstepcounter{figure}\label{fig:lla}
\includepdf[pages=1,fitpaper=true,pagecommand={\thispagestyle{empty}}]{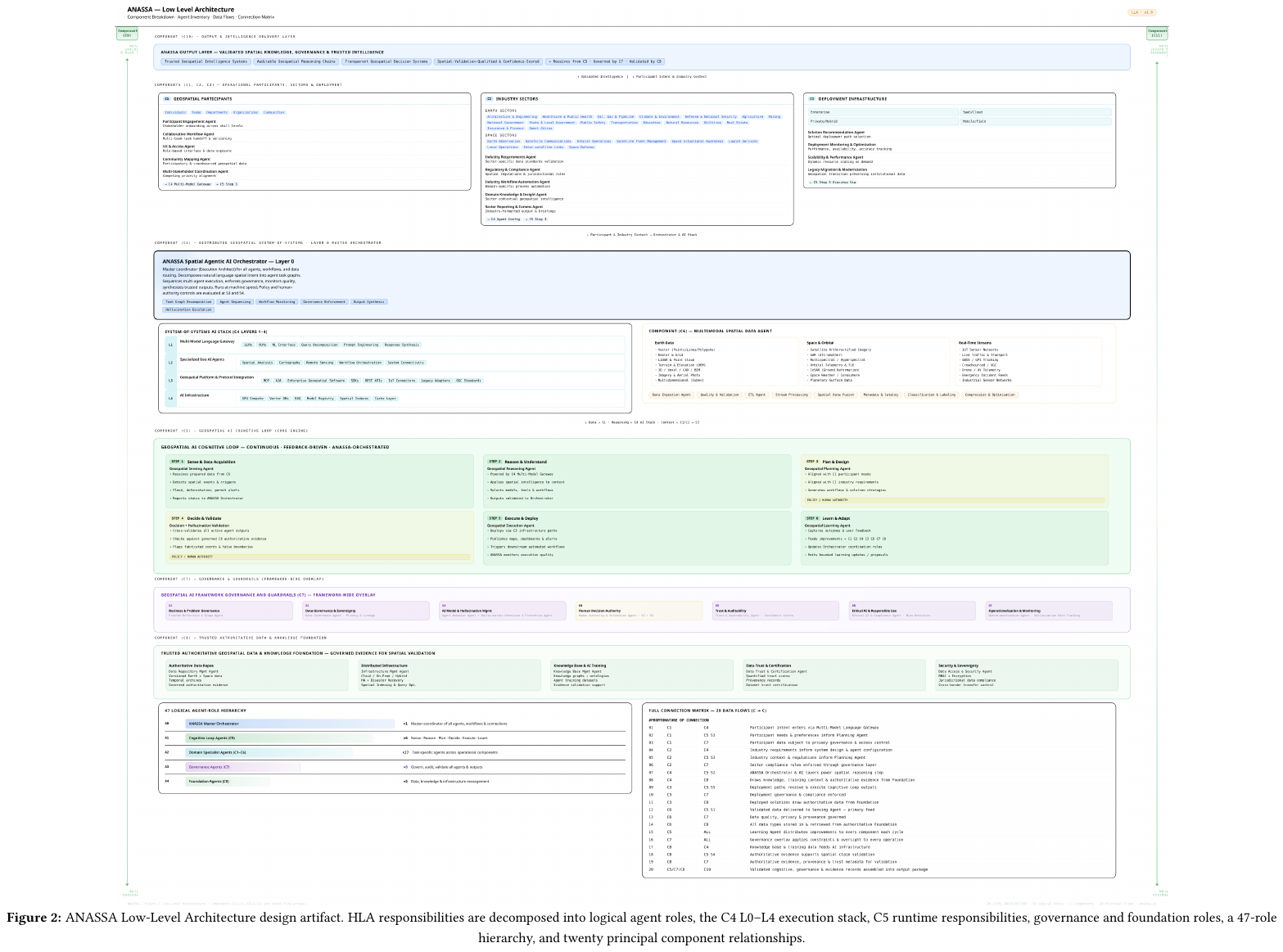}
\clearpage
\twocolumn

\begin{table*}[t]
\caption{Inventory of 47 logical roles mapped to HLA responsibility owners. The count is not a requirement for 47 runtime processes.}
\label{tab:roleinventory}
\scriptsize
\begin{tabular}{p{0.06\textwidth}p{0.08\textwidth}p{0.10\textwidth}p{0.67\textwidth}}
\toprule
\textbf{Tier} & \textbf{Owner} & \textbf{Count} & \textbf{Logical roles} \\
\midrule
A0 & C4-L0 & 1 & ANASSA Spatial Agentic AI Orchestrator \\
A1 & C5 & 6 & Geospatial Sensing; Reasoning; Planning; Decision/Spatial Validation; Execution; Learning \\
A2 & C1 & 5 & Participant Engagement; Collaborative Workflow; UX and Access; Community Mapping; Multi-Stakeholder Coordination \\
A2 & C2 & 5 & Industry Requirements; Regulatory and Compliance; Industry Workflow Automation; Domain Knowledge and Insight; Sector Reporting and Communications \\
A2 & C3 & 4 & Solution Recommendation; Deployment Monitoring and Optimization; Scalability and Performance; Legacy Migration and Modernization \\
A2 & C4-L2 & 5 & Spatial Analysis; Cartography; Remote Sensing; Workflow Orchestration; System Connectivity \\
A2 & C6 & 8 & Data Ingestion; Data Quality and Validation; ETL/Transformation; Stream Processing; Spatial Data Fusion; Metadata and Catalog; Classification and Labeling; Compression and Optimization \\
A3 & C7 & 8 & Problem Definition and Scope; Data Governance; Agent Behavior; Hallucination Detection and Prevention; Human Authority and Validation; Trust and Auditability; Ethical AI and Compliance; Operationalization and Monitoring \\
A4 & C8 & 5 & Data Repository Management; Infrastructure Management; Knowledge Base Management; Data Trust and Certification; Data Access and Security \\
\midrule
 &  & \textbf{47} & \textbf{Total logical roles} \\
\bottomrule
\end{tabular}
\end{table*}

C9 and C11 remain process-flow functions and C10 remains an output-packaging responsibility; they are represented through contracts and runtime records rather than counted as separate logical agents. The eight C7 logical roles refine rather than replace the governance domains inherited from the foundational framework: at the LLA, governance responsibilities are decomposed where separate policy scope, observable state, or validation behavior is required.

The registry also records permitted tools, credentials, expected latency class, resource requirements, and governance scope. A selection decision can therefore be explained as a match between task requirements and registered capabilities.

If no compatible capability is available, the task does not silently degrade. C4 returns a typed state such as \emph{capability-unavailable}, \emph{unsupported-data-type}, \emph{authorization-denied}, or \emph{constraint-conflict}. The planner may then recompose the graph, request clarification, or escalate according to policy.

\subsection{Runtime State Model}

The LLA defines a compact runtime state model for every task. A normal path is:

\begin{quote}\scriptsize
\texttt{RECEIVED} $\rightarrow$ \texttt{PRECONDITION\_CHECK} $\rightarrow$ \texttt{READY}\\
$\rightarrow$ \texttt{SCHEDULED} $\rightarrow$ \texttt{RUNNING} $\rightarrow$ \texttt{VALIDATING}\\
$\rightarrow$ \texttt{COMPLETED}.
\end{quote}

\begin{table}[h]
\caption{Additional ANASSA task runtime states.}
\label{tab:runtimestates}
\centering\scriptsize
\begin{tabularx}{\columnwidth}{p{0.23\columnwidth}>{\raggedright\arraybackslash\ttfamily}X}
Waiting & WAITING\_FOR\_DATA; WAITING\_FOR\_POLICY; WAITING\_FOR\_HUMAN \\
Recovery / terminal & RETRYABLE\_FAILURE; REPLANNING; REJECTED; ESCALATED; TERMINATED \\
\end{tabularx}
\end{table}
Every transition records the cause, the actor or policy that triggered it, a timestamp, the graph revision, and relevant evidence. This makes recovery behavior observable instead of burying it in model-generated reasoning text.

\section{Inter-Component Contracts and Runtime Controls}

\subsection{Canonical Contract Schema}

The foundational ANASSA specification defined architecture-level contract contents and typed failure states for reproducible interfaces~\cite{papantoniou2026anassa}. The LLA refines those elements into the canonical schema (Table~\ref{tab:contract}). The schema is intentionally logical; JSON, protocol buffers, database rows, event envelopes, or other encodings may implement it.

\begin{table*}[t]
\caption{Canonical ANASSA inter-component contract fields.}
\label{tab:contract}
\small
\begin{tabular}{p{0.17\textwidth}p{0.25\textwidth}p{0.50\textwidth}}
\toprule
\textbf{Group} & \textbf{Representative fields} & \textbf{Purpose} \\
\midrule
Identity & contract ID, version, cycle ID, task ID & Binds the exchange to a workflow and schema version \\
Source / destination & component, agent/role, capability & Makes responsibility and routing explicit \\
Spatial context & CRS/datum, units, axis order, geometry type, extent, resolution/scale & Prevents spatial semantics from being implicit \\
Temporal context & observation time, effective interval, freshness/staleness rule & Prevents stale or temporally incompatible reasoning \\
Payload & data type, schema, artifact pointer, semantic type & Defines what is exchanged and how it is interpreted \\
Provenance & source IDs/versions, authority, lineage, transformation history & Supports evidence tracing and reproducibility \\
Execution & tool/capability, parameters, bounds, credential scope, timeout, retry, idempotency & Constrains runtime invocation \\
Validation & validation policy, tolerances, evidence requirements, uncertainty threshold & Defines acceptance criteria before downstream use \\
Governance & risk class, policy IDs, residency/privacy constraints, human-authorization rule & Makes C7 constraints machine-evaluable \\
Response & status, output artifact, evidence bundle, confidence/uncertainty, failure code & Returns both result and machine-readable execution state \\
\bottomrule
\end{tabular}
\end{table*}

A contract is accepted only when preconditions are satisfied. This is important for geospatial systems because a syntactically valid payload may still be operationally invalid: geometry may use the wrong axis order, a raster may be stale, a route may cross a jurisdiction not covered by the policy, or a positional-accuracy claim may exceed the source evidence.

\begin{table}[t]
\caption{Illustrative contract instance used in the worked trace.}
\label{tab:contractinstance}
\scriptsize
\begin{tabularx}{\columnwidth}{p{0.28\columnwidth}X}
\toprule
\textbf{Field} & \textbf{Example value} \\
\midrule
contract / task & K-04 / T-03b \\
source $\rightarrow$ destination & C6 Data Transformation $\rightarrow$ C5-S1 \\
semantic payload & Flood-warning polygon bundle \\
spatial context & EPSG:4326, polygon, extent + source resolution \\
temporal context & observation/effective time + freshness rule \\
provenance & source IDs, versions, transformation from original CRS \\
validation policy & CRS resolvable; geometry valid; evidence not expired \\
governance & analysis-only; no external actuation; policy P-07 \\
response & ACCEPTED + artifact pointer + execution/evidence record \\
\bottomrule
\end{tabularx}
\end{table}

\subsection{Logical Interaction Patterns}

To keep interface semantics independent of transport technology, the LLA defines a compact set of logical interaction patterns:

\begin{itemize}[leftmargin=*,itemsep=2pt]
  \item \textbf{AGENT-INVOKE:} synchronous request/response mediated by the Orchestrator;
  \item \textbf{EVENT-PUB:} asynchronous publication to a declared subscriber set;
  \item \textbf{DATA-READ/WRITE:} versioned access to C8 or governed operational stores;
  \item \textbf{STREAM:} windowed real-time ingest for telemetry and continuously updated observations;
  \item \textbf{POLICY-EVAL:} C7 evaluation returning permit, deny, or permit-with-conditions;
  \item \textbf{HITL-GATE:} a blocking or time-bounded human-authorization request when policy requires it;
  \item \textbf{OUTPUT-ATTEST:} evidence assembly and attestation used by C10 to publish a validation-qualified output package.
\end{itemize}

This vocabulary separates semantic interface types from transport choices. For example, \texttt{EVENT-PUB} may be implemented by a message broker, webhook, event bus, or database notification without changing the logical contract.

\subsection{Execution Controls}

Runtime execution is bounded by tool allowlists, parameter ranges, credential/data-access scope, timeout and retry limits, idempotency expectations, error classification, and rollback or compensating-action rules. The Orchestrator may dispatch only tasks whose preconditions and policy checks are satisfied. A downstream component receives both the produced artifact and its execution record so that later reasoning is grounded in observed execution rather than assumed success.

For side-effecting operations, the contract must also indicate whether the operation is reversible, repeatable, compensatable, or non-idempotent. This becomes especially important when S5 triggers downstream enterprise workflows rather than merely producing maps or analytical layers.

\subsection{Failure Taxonomy and Recovery}

Runtime failures are grouped into five classes. \textbf{Input failures} include schema, CRS, topology, metadata, freshness, or missing-data defects. \textbf{Capability failures} include unavailable tools, unsupported operations, resource constraints, or incompatible models. \textbf{Execution failures} include parameter errors, timeout, downstream-service failure, and partial completion. \textbf{Validation failures} include insufficient evidence, authoritative-source conflict, tolerance violation, and rejected spatial claims. \textbf{Governance failures} include authorization denial, residency/privacy conflict, unapproved risk, or human rejection.

Recovery follows an evidence-driven sequence: classify the observed failure; determine whether a bounded retry is permitted; if not, determine whether the task graph can be replanned with different capabilities or parameters; if not, evaluate whether human adjudication is required; otherwise terminate with a typed failure. The key architectural rule is that replanning is triggered by recorded runtime evidence, not by unconstrained model speculation.

\section{Spatial Validation, Governance, and Provenance}

\subsection{Spatial Validation Pipeline}

S4 operationalizes the spatial-validation requirements established in the foundational ANASSA specification. Candidate outputs are evaluated for coordinate-reference compatibility, geometry/topology, positional accuracy, scale and resolution, temporal validity, uncertainty, jurisdiction, source authority, and conflicting authoritative evidence~\cite{papantoniou2026anassa,iso19157}. These checks can be performed by different services, but the result must be assembled into one validation record before an output is released or a consequential downstream action is authorized.

The LLA distinguishes \emph{claim validation} from \emph{artifact validation}. Artifact validation asks whether the dataset, geometry, map, route, or model output satisfies structural and quality requirements. Claim validation asks whether a specific spatial assertion---for example, a boundary, coordinate, containment relation, risk intersection, or location-dependent recommendation---is supported by admissible evidence. This distinction allows a technically valid dataset to be rejected for a particular decision context when its authority, age, accuracy, or jurisdiction is insufficient.

\subsection{Evidence States}

Consistent with the foundational ANASSA specification, the validation result is not binary. A claim may be \emph{validated}, \emph{validated with uncertainty}, \emph{conflicting authoritative evidence}, \emph{insufficient evidence}, or \emph{rejected}~\cite{papantoniou2026anassa}. Each state carries the checks performed, evidence identifiers and versions, applicable tolerances, uncertainty information, and any human adjudication.

This state model matters for autonomous workflows because ``unknown'' and ``conflicting'' are operationally different from ``false.'' A governed agentic system must be able to stop, escalate, or communicate uncertainty without manufacturing a definitive spatial answer.

\subsection{Risk-Adaptive Human Oversight and Authorization}

The C7 Human Authority and Validation role evaluates whether S3 or S4 requires explicit human authorization. A policy may consider consequence level, output type, jurisdiction, data sensitivity, evidence conflict, uncertainty, financial or safety impact, and whether the workflow is acting on external systems. The result is one of three broad modes: \emph{auto-permit within policy}, \emph{permit with conditions}, or \emph{human authorization required}. The human record includes approver identity/role, timestamp, decision, optional modification, evidence viewed, and the task-graph revision being authorized.

This approach preserves accountable human authority while allowing policy-approved low-risk execution to proceed without inserting a universal manual approval step, consistent with human-AI interaction guidance emphasizing appropriate control, visibility, and recoverability~\cite{amershi2019guidelines}.

\subsection{Provenance Record}

Each completed cycle produces a first-class provenance record containing the objective and C1/C2 context; data, model, and tool versions; task-graph revisions; parameters and intermediate artifacts; execution and validation states; retries and failures; policy and human decisions; learning updates; and final disposition. The record supports decision reconstruction, repeated-run comparison, and attribution of divergence to data, model, tool, or policy changes.

\section{Architecture Integration and Principal Flows}

\subsection{Principal Component Flows}

The LLA consolidates twenty principal component relationships (Table~\ref{tab:matrix}). The matrix does not prohibit additional implementation-specific flows. Its purpose is to identify the architectural pathways that must remain visible across deployments.

\begin{table*}[t]
\caption{ANASSA principal connection matrix.}
\label{tab:matrix}
\scriptsize
\begin{tabular}{c c c p{0.67\textwidth}}
\toprule
\textbf{No.} & \textbf{From} & \textbf{To} & \textbf{Nature of connection} \\
\midrule
1 & C1 & C4 & Participant intent enters through the Multi-Model Language Gateway \\
2 & C1 & C5-S3 & Participant needs and preferences inform Planning Agent context \\
3 & C1 & C7 & Participant data is subject to privacy, authorization, and access-control policy \\
4 & C2 & C4 & Industry requirements inform agent configuration and capability selection \\
5 & C2 & C5-S3 & Industry context and regulations constrain planning \\
6 & C2 & C7 & Sector compliance rules are enforced through governance \\
7 & C4 & C5-S2 & Orchestrator and AI layers provide reasoning capabilities \\
8 & C4 & C8 & C4 draws knowledge, training context, and authoritative evidence from the foundation \\
9 & C5-S5 & C3 & Approved Cognitive Loop outputs are delivered to deployment paths for execution \\
10 & C3 & C7 & Deployment governance and compliance are evaluated \\
11 & C3 & C8 & Deployed solutions may read governed authoritative data \\
12 & C6 & C5-S1 & Prepared and validated data is delivered to the Sensing Agent \\
13 & C6 & C7 & Data quality, privacy, provenance, and sovereignty are governed \\
14 & C6 & C8 & Data types are stored in or retrieved from governed repositories \\
15 & C5-S6 & Permitted subscribers & S6 distributes governed learning updates only to authorized components after a cycle \\
16 & C7 & ALL & Governance constraints and oversight apply across operations \\
17 & C8 & C4 & Knowledge, evidence, and training context feed the AI infrastructure \\
18 & C8 & C5-S4 & Authoritative evidence supports spatial claim validation \\
19 & C8 & C7 & C8 supplies provenance, trust, authority, and validation evidence \\
20 & C5/C7/C8 & C10 & Validated cognitive, governance, and evidence records are assembled into the released output package \\
\bottomrule
\end{tabular}
\end{table*}

\section{DSR Demonstration and Design Evaluation}\label{sec:dsreval}

\subsection{Synthetic End-to-End Architecture Demonstration}

To check that the LLA mechanisms close into one coherent execution path, Table~\ref{tab:trace} instantiates a synthetic, non-performance trace for the objective: \emph{identify facilities within 5 km of a current flood-warning polygon and release only spatial claims supported by admissible evidence}. The trace is intentionally small; its purpose is to exercise task/contract identity, a spatial precondition failure, graph revision, policy evaluation, evidence-based validation, C10 packaging, and provenance closure.

\begin{table*}[t]
\caption{Synthetic architecture trace demonstrating failure, recovery, validation, and provenance. Times are logical offsets, not measured latency.}
\label{tab:trace}
\scriptsize
\begin{tabular}{p{0.04\textwidth}p{0.05\textwidth}p{0.08\textwidth}p{0.15\textwidth}p{0.18\textwidth}p{0.29\textwidth}}
\toprule
\textbf{Time} & \textbf{Task} & \textbf{Owner} & \textbf{State} & \textbf{Contract / evidence} & \textbf{Observed event and architectural response} \\
\midrule
$\tau_0$ & T-01 & C1/C4 & RECEIVED & K-01 intent & Objective, role, output type, and policy context normalized; graph $G_0$ created. \\
$\tau_1$ & T-02 & C6/C8 & COMPLETED & K-02 data read & Facility points and flood-warning polygon resolved with versions, timestamps, and authority metadata. \\
$\tau_2$ & T-03 & C6 & \shortstack{PRECONDITION\\CHECK} & K-03 spatial bundle & Contract rejected because the two inputs use incompatible CRS definitions for the requested distance computation. \\
$\tau_3$ & T-03 & C4 & \shortstack{RETRYABLE\\FAILURE} & \shortstack{failure=\\CRS\_MISMATCH} & Failure evidence is recorded; unconstrained execution is blocked. \\
$\tau_4$ & T-03b & C4/C6 & REPLANNING & $G_1$, K-04 & Orchestrator inserts an explicit governed transformation task; original $G_0$ remains immutable. \\
$\tau_5$ & T-03b & C6 & COMPLETED & K-04 accepted & Inputs transformed to a distance-suitable CRS; transformation lineage retained. \\
$\tau_6$ & T-04 & C5-S3/C7 & READY & policy P-07 & Plan is classified analysis-only and auto-permitted within policy; no human block is required. \\
$\tau_7$ & T-05 & C5-S4/C8 & VALIDATING & evidence bundle E-05 & Geometry, temporal validity, authority, 5-km distance relation, and source applicability are checked. \\
$\tau_8$ & T-05 & C5-S4 & COMPLETED & \shortstack{VALIDATED WITH\\UNCERTAINTY} & Claim is supported; positional uncertainty inherited from the facility dataset is preserved in the verdict. \\
$\tau_9$ & T-06 & C10/C11 & COMPLETED & output O-06 + provenance R-06 & C10 packages the qualified claim and decision record; C11 closes the cycle with graph, contract, failure, policy, evidence, and output links. \\
\bottomrule
\end{tabular}
\end{table*}

The worked trace is an architecture-level consistency demonstration, not an empirical result. It shows that a spatial mismatch becomes a typed contract failure rather than a hidden prompt correction; replanning creates a new DAG revision; policy controls can permit a low-consequence analytical path without mandatory human latency; S4 retains uncertainty instead of forcing a binary verdict; and the final C10 package can be reconstructed from the cycle record.

\subsection{Analytical Evaluation Criteria}

The design artifact is evaluated against four criteria established in the DSR methodology.

\begin{table*}[t]
\caption{Analytical evaluation of the ANASSA design artifact.}
\label{tab:eval}
\small
\begin{tabular}{p{0.10\textwidth}p{0.23\textwidth}p{0.43\textwidth}p{0.09\textwidth}}
\toprule
\textbf{Criterion} & \textbf{Evaluation question} & \textbf{Evidence in the design artifact} & \textbf{Result} \\
\midrule
E1 Requirements coverage & Do inherited, functional, and non-functional requirements have identifiable design owners and mechanisms? & DR1--DR8 map to DP1--DP8, HLA owners, LLA mechanisms, and observable records; FR/NFR responsibilities are represented across C1--C11 and runtime controls. & Covered \\
E2 Internal consistency & Are responsibility, direction, policy, evidence, and failure semantics mutually consistent? & C7 owns governance, C8 owns governed authoritative evidence, S4 owns validation, C10 packages released results; twenty principal flows have explicit direction; failure states terminate in retry, replan, escalation, compensating action, or termination. & Consistent by design review \\
E3 End-to-end closure & Can a representative objective traverse the architecture without relying on hidden corrective behavior? & Table~\ref{tab:trace} demonstrates task identity, contract rejection, CRS failure, graph revision, policy decision, spatial validation, uncertainty retention, output packaging, and provenance closure. & Demonstrated \\
E4 Technology independence & Can logical orchestration remain stable while implementation choices vary? & DP3 separates capabilities from providers; C4-L3/L4 are implementation surfaces; contracts and task semantics remain above specific models, geospatial platforms, clouds, and protocols. & Supported architecturally \\
\bottomrule
\end{tabular}
\end{table*}

This evaluation establishes design completeness and internal coherence, not operational superiority. Claims of execution accuracy, latency, cost, throughput, resilience, or scalability require an implemented system and controlled comparison.

\subsection{Reference Implementation Readiness}

A conforming reference implementation does not need to reproduce every diagram element as a separate service. It must expose enough state to verify the design artifact. At minimum, an implementation should provide: (1) stable identifiers for cycles, tasks, contracts, roles, and graph revisions; (2) explicit spatial/temporal context on task inputs; (3) a capability registry; (4) machine-readable policy decisions; (5) typed execution and validation states; (6) evidence/provenance pointers; (7) observable retry, replanning, and escalation events; and (8) a final C10 decision record linking the output to its supporting execution path.

Technology independence should be tested at component boundaries rather than by requiring a single platform. C4-L3 is therefore a critical implementation surface: the same logical task and contract should be bindable to alternative geospatial services or tools while preserving the orchestration semantics above the adapter layer.

\section{Discussion}

\subsection{Generalizable Design Knowledge}

The primary design contribution is not the visual form of the HLA or LLA. It is the set of reusable design decisions that connect geospatial intent to governed execution: geospatial context travels with the task; intent is separated from execution; capabilities are resolved independently of vendors; cross-boundary exchanges are typed; validation is part of the runtime path; autonomy is policy- and risk-adaptive; failures become evidence for bounded recovery; and released decisions remain reconstructable. These principles can guide implementations that differ substantially in models, tools, platforms, and deployment topology.
\enlargethispage{4\baselineskip}

\subsection{System Architecture Implications}

The architecture follows a product/system design discipline in which user and organizational objectives are translated into requirements before implementation technologies are selected. C1/C2 express user, stakeholder, sector, and policy context; C4/C5 translate those objectives into governed execution; C6/C8 provide data and evidence; C7 constrains autonomy; C3 isolates deployment choices; and C10/C11 make outputs and decision history consumable. This separation is important because a successful geospatial agentic system must optimize more than model quality: it must also support interoperability, observability, failure recovery, security, accountability, and change over time.

\subsection{Relationship to the Foundational ANASSA Framework}

The foundational ANASSA study established the framework-level architecture, component boundaries, architectural requirements, and evaluation variables~\cite{papantoniou2026anassa}. The present study extends that baseline through systematic HLA-to-LLA refinement, explicit logical roles, versioned task graphs, capability resolution, canonical contracts, runtime states, recovery behavior, and analytical design evaluation. The next research stage should instantiate a minimal vertical slice and evaluate architecture conformance before broader domain benchmarking. This separation preserves design quality and empirical performance as distinct research questions.

\section{Limitations, Threats to Validity, and Future Work}

The design remains an architecture artifact rather than an implemented production system. It does not establish that 47 logical roles are optimal, that one task-graph representation will suit every geospatial workload, or that every workflow needs all twenty principal flows. The role inventory may be consolidated or extended during implementation provided that responsibility and audit boundaries remain observable. Likewise, the interaction-pattern vocabulary is a design abstraction rather than a requirement for a specific transport protocol.

The LLA also does not prescribe distributed-runtime consistency semantics such as exactly-once versus at-least-once delivery, global ordering, backpressure, leader election, distributed locking, cancellation propagation, or recovery after an Orchestrator process failure. Those properties depend on the selected transport and runtime. A reference implementation must therefore declare its delivery, ordering, idempotency, checkpoint, and partial-failure guarantees and demonstrate that they preserve the logical task/contract semantics defined here.

Because the evaluation is analytical and author-conducted, it can establish traceability and internal coherence but not external validity, operational effectiveness, or superiority over alternative architectures. The synthetic trace exercises selected failure and governance paths rather than the full space of geospatial workloads, and the 47-role inventory has not yet been empirically optimized. These limitations motivate implementation-based conformance testing and comparative evaluation.

Future work should proceed in three stages. First, implement a minimal vertical slice of C4/C5/C7/C8/C10/C11 and measure architecture conformance through contract completeness, trace reconstruction, typed-failure capture, graph-revision correctness, policy-decision observability, and evidence-chain completeness. Second, demonstrate DP3 deployment independence by binding the same logical task/contract to more than one geospatial execution adapter. Third, benchmark reasoning, tool sequencing, spatial validation, failure recovery, human escalation, reproducibility, and scalability against geospatial-agent baselines such as GeoAgentBench and related systems~\cite{liang2026geoagenticrag,yu2026geoagentbench}.

\section{Conclusion}

This paper presents an ANASSA HLA-to-LLA refinement design for spatial agentic AI. Starting from the framework-level architecture established in the foundational ANASSA study~\cite{papantoniou2026anassa}, it derives implementation-facing requirements, formalizes eight LLA-focused design principles, interprets the inherited structure as an HLA responsibility model, and systematically refines those responsibilities into the LLA. The resulting artifact includes a five-tier hierarchy of 47 logical roles, versioned task graphs, capability resolution, typed contracts, observable runtime states, risk-adaptive governance, spatial-validation pathways, evidence-driven recovery, provenance records, and principal component flows.

The central design proposition is that governed geospatial agentic AI should not be organized around a particular foundation model or geospatial product. It should be organized around stable responsibilities, replaceable capabilities, explicit contracts, observable state, governed evidence, and reconstructable decisions.

\bibliographystyle{ACM-Reference-Format}
\bibliography{references}

\end{document}